\documentclass{article}
\usepackage{microtype}
\usepackage{graphicx}
\usepackage{subfigure}
\usepackage{booktabs}

\usepackage{hyperref}

\usepackage[accepted]{icml2024}

\usepackage{amsmath}
\usepackage{amssymb}
\usepackage{mathtools}
\usepackage{amsthm}

\usepackage[capitalize,noabbrev]{cleveref}

\theoremstyle{plain}

\theoremstyle{definition}

\theoremstyle{remark}

\usepackage[textsize=tiny]{todonotes}

\icmltitlerunning{TICA Energy Matching for Machine-Learned Molecular Dynamics}

\begin{document}

\twocolumn[
\icmltitle{TICA-Based Free Energy Matching for Machine-Learned Molecular Dynamics}



\icmlsetsymbol{equal}{*}

\begin{icmlauthorlist}
\icmlauthor{Alexander Aghili}{ucsc,equal}
\icmlauthor{Andy Bruce}{ucsc,equal}
\icmlauthor{Daniel Sabo}{ucsc}
\icmlauthor{Razvan Marinescu}{ucsc}
\end{icmlauthorlist}

\icmlaffiliation{ucsc}{Baskin Engineering, University of California - Santa Cruz, Santa Cruz, United States}

\icmlcorrespondingauthor{Alexander Aghili}{awaghili@ucsc.edu}
\icmlcorrespondingauthor{Andy Bruce}{acbruce@ucsc.edu}

\icmlkeywords{Machine Learning, ICML, Biomolecular, Biomed, Molecular Dynamics, Thermodynamics, Statistical Mechanics, Proteins}

\vskip 0.3in
]



 
\printAffiliationsAndNotice{\icmlEqualContribution} 

\begin{abstract}

Molecular dynamics (MD) simulations provide atomistic insight into biomolecular systems but are often limited by high computational costs required to access long timescales. Coarse-grained machine learning models offer a promising avenue for accelerating sampling, yet conventional force matching approaches often fail to capture the full thermodynamic landscape as fitting a model on the gradient may not fit the absolute differences between low-energy conformational states. In this work, we incorporate a complementary energy matching term into the loss function. We evaluate our framework on the Chignolin protein using the CGSchNet model, systematically varying the weight of the energy loss term. While energy matching did not yield statistically significant improvements in accuracy, it revealed distinct tendencies in how models generalize the free energy surface. Our results suggest future opportunities to enhance coarse-grained modeling through improved energy estimation techniques and multi-modal loss formulations.

\end{abstract}

\section{Introduction}
\label{intro}

Molecular dynamics (MD) simulations are a cornerstone of computational biophysics, offering atomistic insights into protein folding, conformational transitions, and molecular interactions. However, the fine temporal and spatial resolution required to resolve these events imposes severe computational costs. Due to the high dimensionality of molecular systems and the presence of stiff intramolecular forces, MD simulations typically use femtosecond-scale integration time steps, necessitating billions of steps to observe biologically relevant timescales. This bottleneck has prompted significant interest in alternative strategies that accelerate sampling without compromising physical realism \cite{Kidder_2021}.

Recent advances in machine learning have enabled the construction of coarse-grained potentials that approximate free energy surfaces from atomistic data. Most models rely on force matching, which fits the gradients of the free energy surface to reproduce local forces. While effective for capturing short-timescale dynamics, this approach can miss global thermodynamic structure since it ignores the absolute energy values and may poorly distinguish between metastable states \cite{fu2023forcesenoughbenchmarkcritical, 10.1063/5.0124538}.

To address these limitations, we incorporated energy matching into the learning objective by leveraging the Boltzmann relation, which links free energies to probability densities over reduced coordinates. This approach enabled us to supervise the model not only with force data but also with approximate free energy estimates. While our current implementation of energy matching did not yield a statistically significant performance improvement, the results highlight promising directions for enhancing the model. These insights offer a foundation for refining our methodology and potentially achieving more accurate and robust outcomes in future iterations.

\section{Time-lagged Independent Component Analysis Background}
\label{tica}
Time-lagged Independent Component Analysis (TICA) is a linear dimensionality reduction technique that isolates the slowest dynamical modes in time-series data \cite{doi:10.1021/acs.jctc.5b00743}. These modes often correspond to biologically meaningful transitions, such as folding, conformational switching, or binding events. By constructing a new coordinate system where components are both uncorrelated and optimized for long-time autocorrelation, TICA filters out fast, thermally driven fluctuations and emphasizes the kinetically dominant directions in configuration space. The TICA process begins with a set of high-dimensional features, such as backbone dihedrals, contact maps, or pairwise distances, represented as time-ordered vectors \(\mathbf{x}(t)\). Unlike PCA, which solves the simple eigenvalue decomposition
\[
C_0\vec{v} = \lambda \vec{v}
\]
where \(c_{ij} = \text{Cov}(x_i(t), x_j(t))\), TICA computes the time-lagged and instantaneous covariance matrices, then solves the generalized eigenvalue problem
\[
C'(\tau) \vec{v} = \lambda C_0 \vec{v}
\]
where \(C'(\tau)\) is a time lagged covariance matrix where
\[
c'_{ij}(\tau) = \text{Cov}(x_i(t), x_j(t + \tau))
\]

Due to sampling variance, the empirical covariance $\text{Cov}(x_i(t), x_j(t + \tau))$ may differ from $\text{Cov}(x_i(t + \tau), x_j(t))$, even though they are theoretically equal in the absence of noise. To address this, the covariance is typically symmetrized with 
\begin{align*}
  c'_{ij}(\tau) = \frac{1}{2} &\text{Cov}(x_i(t), x_j(t + \tau)) +\\ \frac{1}{2}&\text{Cov}(x_i(t + \tau), x_j(t))
\end{align*}
to ensure that the resulting matrix is Hermitian. This symmetrization guarantees real eigenvalues and eigenvectors, which is often desirable in practical applications.

The resulting eigenvectors \(\mathbf{v}_i\) define directions in the input space with maximal time-autocorrelation, and the projections \(y_i(t) = \mathbf{v}_i^{-1}x(t)\) serve as the reduced coordinates.

Once a trajectory is projected into the TICA space, the probability density over this space can be estimated using a KDE or histogram. From this, the Gibbs free energy $G$ can be computed up to a constant by applying the Boltzmann relation:
\[
    G(\mathbf{y}) = -k_BT\log(P(\mathbf{y}))
\]
where 
\(P(\mathbf{y})\) is the estimated density in TICA space and \(-k_BT\) is the thermal energy. This transformation reveals the free energy surface, a map where valleys correspond to metastable conformations and barriers suggest transition states. 
\section{Force matching}
Force matching on a continuous coarse grained function \(f \in \mathbb{R}^{3N} \rightarrow \mathbb{R}^{3n} \) usually defines the loss to be the gradient of the free energy \cite{https://doi.org/10.1002/cphc.200400669}
\begin{multline}
  \label{force_loss}
  L_{\text{force}}(\theta) = \frac{1}{N} \sum_{i=1}^N  \Big\lVert \nabla_{\vec{R}} U(f(\vec{r}))\Big( \mathbf{B}(\vec{r}_i) \nabla_{\vec{r}} E(\vec{r}_i) \\
  - k_B T \mathbf{B}(\vec{r}_i) \nabla_{\vec{r}} \Big) \Big\rVert_{2}
\end{multline}

Langevin sampling will by nature sample lower energy states, meaning that the high energy transitions between energy minima may not be well sampled. Due to only matching the gradient, the absolute free energy levels between low energy conformational states may not be accurate.

\section{Energy Matching Theory}
\label{energy_matching}

The Boltzmann distribution describes the probability of a configuration $R$ in the canonical ensemble:
\[
P(R) = \frac{e^{-\frac{G(R)}{k_B T}}}{Z}
\]
where $G(R)$ is the free energy, $k_B$ is the Boltzmann constant, $T$ is the temperature, and $Z$ is the partition function.

By inverting this relationship, we obtain an expression for the free energy:
\[
G(R) = -k_B T \ln(P(R)) + C
\]
where $C = -k_B T \ln(Z)$ is an additive constant.

Assuming that the configuration vector $R \in \mathbb{R}^{3n}$ can be transformed into a set of statistically independent components $K \in \mathbb{R}^m$, such that $\text{Cov}(K_i, K_j) = 0$ for $i \ne j$, the free energy can be decomposed as:
\begin{align}
G(R) &= -k_B T \ln(P(K)) + C \\
     &= -k_B T \sum_{i=1}^m \ln(P(K_i)) + C
\end{align}
This formulation allows the free energy to be estimated up to an additive constant based on the probability distribution of the independent components $K_i$.

To incorporate this into model training, we introduce an energy-based loss term alongside the force matching loss from Section~\ref{force_loss}. The total loss function becomes:
\begin{align*}
    L(\theta) = &\lambda_{\text{force}} L_{\text{force}}(\theta) \\& + \lambda_{\text{energy}} \frac{1}{N} \sum_{i=1}^N \left[ U(\theta, \vec{R}_i) - G(R_i) + C \right]
\end{align*}
Here, $U(\theta, \vec{R}_i)$ is the predicted energy for configuration $\vec{R}_i$, and $C$ is a protein-specific constant allowing for comparison up to an additive offset. The hyperparameters $\lambda_{\text{force}}$ and $\lambda_{\text{energy}}$ control the trade-off between force matching and energy matching in the optimization objective.

\section{Methodology}
\label{methodology}
We used the CGSchnet model \cite{Zhang_2020} with a coarse grain function as a matrix keeping only the carbon alpha atoms of the protein. Our initial data was derived from Majewski et al. \cite{Majewski_2023} at 350K, and resampled from diverse initial conformations at 300K. 

For each TICA component $k \in \{1, \ldots, d\}$, we estimate the marginal probability density function $P_k(y)$ of component values across the entire dataset using either a Gaussian kernel density estimator or a histogram-based approach. The free energy of a trajectory point $y$ along a TICA component is computed using the Boltzmann inversion:
\[
G_k(y) = -k_B T \log P_k(y),
\]
where $k_B$ is the Boltzmann constant and $T$ is the temperature (set to 300 K in our experiments). To obtain the total free energy for a point $\mathbf{y}$ in TICA space, we sum the contributions from each independent component:
\[
G(\mathbf{y}) = \sum_{k=1}^d G_k(y_k) = -k_B T \sum_{k=1}^d \log P_k(y_k).
\]
For each trajectory, we load the precomputed prior energy $E_{\text{prior}}$ and calculate the energy correction $\Delta E = G(\mathbf{y}) - E_{\text{prior}}$. This correction $\Delta E$ is saved and can be used as the scalar target in machine learning models trained to predict absolute free energy differences across protein conformational states.
\begin{figure}[H]
\vskip 0.2in
\begin{center}
\centerline{\includegraphics[width=\columnwidth]{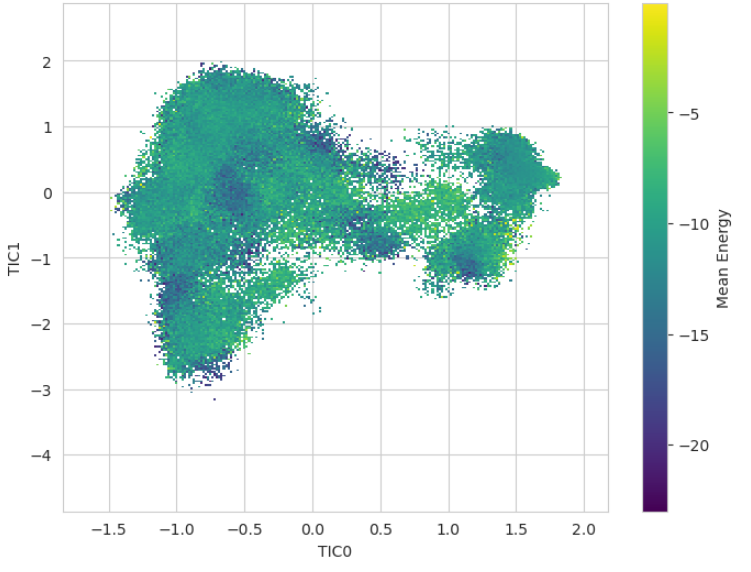}}
\caption{Predicted Mean TICA free energy landscape of Chignolin projected onto TICA space. Energy values were computed by binning TICA data into a 100×100 grid and averaging over each bin.}
\label{meanenergy}
\end{center}
\vskip -0.2in
\end{figure}

\section{Results}
\label{results}

Regions with the greatest density of TICA points are expected to correlate strongly with low-energy conformational states of the system. These areas likely represent thermodynamically stable states, where the system resides for extended periods due to minimal free energy, as inferred from the clustering behavior in TICA space. 

\begin{figure}[hb]
\vskip 0.2in
\begin{center}
\centerline{\includegraphics[width=\columnwidth]{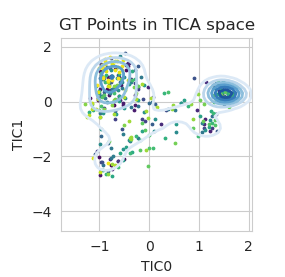}}
\caption{Contour plot illustrating the density of Ground-Truth TICA points, overlaid with strided data points for clarity.}
\label{gttica}
\end{center}
\vskip -0.2in
\end{figure}

We compare the distribution of point density in TICA space with the corresponding TICA energy landscape and observe that the resulting energy landscape is noticeably flatter than the empirical density, suggesting that the marginal energy estimates may underrepresent the depth and sharpness of energy wells as illustrated by \cref{meanenergy} and \cref{gttica}. This raises the possibility that TICA energy may not capture the full extent of the system’s dynamical or thermodynamic structure.

We trained a series of models on the Chignolin protein, varying the values of the weighting parameters \(\lambda_{\text{energy}}\) and \(\lambda_{\text{force}}\), constrained such that \(\lambda_{\text{energy}} + \lambda_{\text{force}} = 1\). We assess performance using the Kullback-Leibler (KL) divergence between the model output and ground-truth data, computed on the first two TICA components. These components are chosen because they account for over 70\% of the variance in the system’s slow dynamics. TICA determines the explained variance of each component based on its associated eigenvalue, making the first two components the most informative for capturing the dominant slow modes.

\begin{figure}[H]
\vskip 0.2in
\begin{center}
\centerline{\includegraphics[width=\columnwidth]{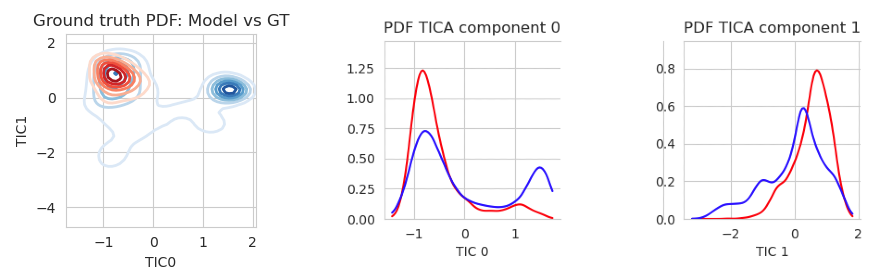}}
\caption{Plot of TICA points with $\lambda_{energy} = 0$ (pure force matching) for inference on a Chignolin trained model (Blue is ground-truth, red is machine learning model). Includes a contour representing the density of the TICA points of TIC 0 and TIC 1, as well as the associated PDFs.}
\label{chignolinforce}
\end{center}
\vskip -0.2in
\end{figure}

\begin{figure}[H]
\vskip 0.2in
\begin{center}
\centerline{\includegraphics[width=\columnwidth]{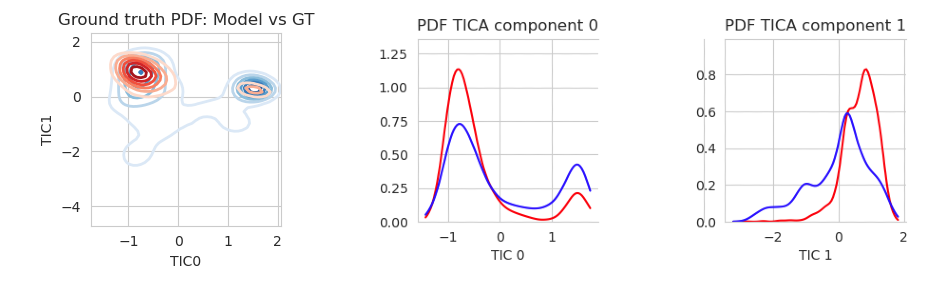}}
\caption{Plot of TICA points with $\lambda_{energy} = 0.01$ for Chignolin model.}
\label{chignolinlow}
\end{center}
\vskip -0.2in
\end{figure}

\begin{figure}[H]
\vskip 0.2in
\begin{center}
\centerline{\includegraphics[width=\columnwidth]{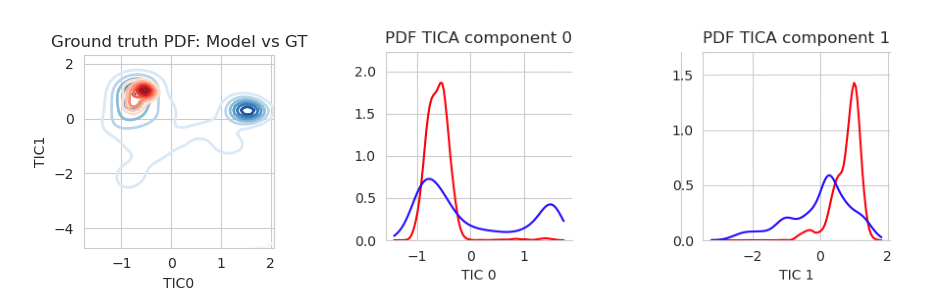}}
\caption{Plot of TICA points with $\lambda_{energy} = 0.8$ for Chignolin model.}
\label{chignolinhigh}
\end{center}
\vskip -0.2in
\end{figure}

Models trained with low values of \(\lambda_{\text{energy}}\) (e.g., 0.01, 0.05, 0.075) when compared to ground truth force matching demonstrated a slight improvement in accurately capturing the free energy surface seen in \cref{chignolinforce} compared to \cref{chignolinlow}. However, the differences observed may be attributable to stochastic variation as there was no statistically significant difference in free energy estimation accuracy. As \(\lambda_{\text{energy}}\) was increased to intermediate and high values (0.1, 0.5, and 1.0), the models exhibited a tendency to overfit to the lowest-energy basins, becoming trapped in deep wells of the potential energy surface as displayed in \cref{chignolinhigh}. 

\section{Future Work}
\label{future_work}
Although our energy matching results did not demonstrate significant improvements with the techniques presented, we believe that further research and alternative methodologies hold promise for effectively learning free energy.

We theorize that more complex proteins, such as Protein G or a3D, which exhibit deeper and more varied energy basins, may present a better testbed for evaluating energy matching techniques. To test this hypothesis, we plan to extend our methodology to such proteins and assess whether the richer energy structures improve learning and generalization. Additionally, we will explore simplified, synthetic benchmark systems with known energy surfaces to better isolate the effects of energy well depth, barrier heights, and statistical dependencies. These controlled environments will help us disentangle model performance from biological variability and guide the refinement of energy-based loss formulations. While the current results highlight challenges, they also underscore important design considerations that will inform our future efforts in learning accurate free energy surfaces.

While we have utilized TICA-based density estimation to approximate free energies, alternative approaches offer complementary insights into the thermodynamics and kinetics of molecular systems. For example, Markov State Models (MSMs) provide a principled method for analyzing long-timescale behavior by discretizing the system's configuration space into metastable microstates and estimating transition probabilities between them over a fixed lag time $\tau$.

MSMs can be constructed by clustering molecular configurations in the TICA-projected space and computing the transition probability matrix $T$ from observed trajectory transitions. The stationary distribution $\pi$ of $T$ then yields an equilibrium probability for each state, from which free energies can be recovered using the Boltzmann relation:
\[
G_i = -k_B T \log(\pi_i)
\]
This approach effectively converts high-dimensional dynamical information into a coarse-grained free energy landscape.

We aim to explore MSM-derived free energies as an alternative to TICA-based density estimates for the energy loss term. This could enhance model robustness, especially for systems with complex metastable behavior, and support a broader range of physical constraints in energy matching.

\bibliography{tica_paper}

\begin{thebibliography}{7}
\providecommand{\natexlab}[1]{#1}
\providecommand{\url}[1]{\texttt{#1}}
\expandafter\ifx\csname urlstyle\endcsname\relax
  \providecommand{\doi}[1]{doi: #1}\else
  \providecommand{\doi}{doi: \begingroup \urlstyle{rm}\Url}\fi

\bibitem[Ciccotti et~al.(2005)Ciccotti, Kapral, and Vanden-Eijnden]{https://doi.org/10.1002/cphc.200400669}
Ciccotti, G., Kapral, R., and Vanden-Eijnden, E.
\newblock Blue moon sampling, vectorial reaction coordinates, and unbiased constrained dynamics.
\newblock \emph{ChemPhysChem}, 6\penalty0 (9):\penalty0 1809--1814, 2005.
\newblock \doi{https://doi.org/10.1002/cphc.200400669}.
\newblock URL \url{https://chemistry-europe.onlinelibrary.wiley.com/doi/abs/10.1002/cphc.200400669}.

\bibitem[Fu et~al.(2023)Fu, Wu, Wang, Xie, Keten, Gomez-Bombarelli, and Jaakkola]{fu2023forcesenoughbenchmarkcritical}
Fu, X., Wu, Z., Wang, W., Xie, T., Keten, S., Gomez-Bombarelli, R., and Jaakkola, T.
\newblock Forces are not enough: Benchmark and critical evaluation for machine learning force fields with molecular simulations, 2023.
\newblock URL \url{https://arxiv.org/abs/2210.07237}.

\bibitem[Kidder et~al.(2021)Kidder, Szukalo, and Noid]{Kidder_2021}
Kidder, K.~M., Szukalo, R.~J., and Noid, W.~G.
\newblock Energetic and entropic considerations for coarse-graining.
\newblock \emph{The European Physical Journal B}, 94\penalty0 (7):\penalty0 2, July 2021.
\newblock ISSN 1434-6036.
\newblock \doi{10.1140/epjb/s10051-021-00153-4}.
\newblock URL \url{https://doi.org/10.1140/epjb/s10051-021-00153-4}.

\bibitem[Majewski et~al.(2023)Majewski, Pérez, Thölke, Doerr, Charron, Giorgino, Husic, Clementi, Noé, and De~Fabritiis]{Majewski_2023}
Majewski, M., Pérez, A., Thölke, P., Doerr, S., Charron, N.~E., Giorgino, T., Husic, B.~E., Clementi, C., Noé, F., and De~Fabritiis, G.
\newblock Machine learning coarse-grained potentials of protein thermodynamics.
\newblock \emph{Nature Communications}, 14\penalty0 (1), September 2023.
\newblock ISSN 2041-1723.
\newblock \doi{10.1038/s41467-023-41343-1}.
\newblock URL \url{http://dx.doi.org/10.1038/s41467-023-41343-1}.

\bibitem[Scherer et~al.(2015)Scherer, Trendelkamp-Schroer, Paul, P{\'e}rez-Hernández, Hoffmann, Plattner, Wehmeyer, Prinz, and No{\'e}]{doi:10.1021/acs.jctc.5b00743}
Scherer, M.~K., Trendelkamp-Schroer, B., Paul, F., P{\'e}rez-Hernández, G., Hoffmann, M., Plattner, N., Wehmeyer, C., Prinz, J.-H., and No{\'e}, F.
\newblock Pyemma 2: A software package for estimation, validation, and analysis of markov models.
\newblock \emph{Journal of Chemical Theory and Computation}, 11\penalty0 (11):\penalty0 5525--5542, 2015.
\newblock \doi{10.1021/acs.jctc.5b00743}.
\newblock URL \url{https://doi.org/10.1021/acs.jctc.5b00743}.
\newblock PMID: 26574340.

\bibitem[Thaler et~al.(2022)Thaler, Stupp, and Zavadlav]{10.1063/5.0124538}
Thaler, S., Stupp, M., and Zavadlav, J.
\newblock Deep coarse-grained potentials via relative entropy minimization.
\newblock \emph{The Journal of Chemical Physics}, 157\penalty0 (24):\penalty0 244103, 12 2022.
\newblock ISSN 0021-9606.
\newblock \doi{10.1063/5.0124538}.
\newblock URL \url{https://doi.org/10.1063/5.0124538}.

\bibitem[Zhang et~al.(2020)Zhang, Han, Wang, Car, and E]{Zhang_2020}
Zhang, L., Han, J., Wang, H., Car, R., and E, W.
\newblock Coarse graining molecular dynamics with graph neural networks.
\newblock \emph{The Journal of Chemical Physics}, 153\penalty0 (19):\penalty0 5--10, November 2020.
\newblock ISSN 1089-7690.
\newblock \doi{10.1063/5.0026133}.
\newblock URL \url{https://doi.org/10.1063/5.0026133}.

\end{thebibliography}
\bibliographystyle{icml2024}

\newpage
\appendix
\onecolumn

\end{document}